\documentclass{article}

\usepackage[accepted]{icml2017}

\usepackage{natbib}
\usepackage[utf8]{inputenc} 
\usepackage[T1]{fontenc}    
\usepackage{hyperref}       
\usepackage{url}            
\usepackage{booktabs}       
\usepackage{amsfonts}       
\usepackage{nicefrac}       
\usepackage{microtype}      
\usepackage[]{todonotes}
\usepackage{subcaption}
\usepackage{adjustbox}
\usepackage{authblk}
\usepackage{amsmath}

\usepackage{tikz}
\usetikzlibrary{backgrounds,fit,decorations.pathreplacing, external}  
\usepackage{pgfplots}
\pgfplotsset{compat=1.13}

\author[1]{Emilio Jorge}
\author[2]{Mikael K{\aa}geb{\"a}ck}
\author[3]{Fredrik D. Johansson}
\author[1]{Emil Gustavsson}

\affil[1]{Fraunhofer-Chalmers Centre \\
G\"{o}teborg, Sweden \\
\texttt{firstname.lastname@fcc.chalmers.se}}
\affil[2]{Computer Science \& Engineering \\
Chalmers University of Technology \\
G\"{o}teborg, Sweden \\
\texttt{kageback@chalmers.se}}
\affil[3]{IMES, Massachusetts Institute of Technology\\
Massachusetts, USA \\
\texttt{ml@fredjo.com}}

\DeclareMathOperator*{\argmax}{argmax}

\begin{document}
\twocolumn[
\icmltitle{Learning to Play \emph{Guess Who?} and Inventing a Grounded Language as a Consequence}

\icmltitlerunning{Learning to Play \emph{Guess Who?} and Inventing a Grounded Language as a Consequence}

\icmlauthor{Emilio Jorge}{emilio.jorge@fcc.chalmers.se}
\icmladdress{Fraunhofer-Chalmers Centre, G\"{o}teborg, Sweden}
\icmlauthor{Mikael K{\aa}geb{\"a}ck}{kageback@chalmers.se}
\icmladdress{Computer Science \& Engineering, Chalmers University of Technology, G\"{o}teborg, Sweden}
\icmlauthor{Fredrik D. Johansson}{ml@fredjo.com}
\icmladdress{IMES, Massachusetts Institute of Technology, Massachusetts, USA}
\icmlauthor{Emil Gustavsson}{emil.Gustavsson@fcc.chalmers.se}
\icmladdress{Fraunhofer-Chalmers Centre, G\"{o}teborg, Sweden}


\vskip 0.3in
]

\begin{abstract}
  Acquiring your first language is an incredible feat and not easily duplicated. Learning to communicate using nothing but a few pictureless books, a corpus, would likely be impossible even for humans. Nevertheless, this is the dominating approach in most \emph{natural language processing} today. As an alternative, we propose the use of situated interactions between agents as a driving force for communication, and the framework of \emph{Deep Recurrent Q-Networks} for evolving a shared language grounded in the provided environment. We task the agents with interactive image search in the form of the game \emph{Guess Who?}. The images from the game provide a non trivial environment for the agents to discuss and a natural grounding for the concepts they decide to encode in their communication. Our experiments show that the agents learn not only to encode physical concepts in their words, i.e. grounding, but also that the agents learn to hold a multi-step dialogue remembering the state of the dialogue from step to step.
\end{abstract}

\section{Introduction}

Human language retains little meaning without interactions with the world it is used to describe, e.g. reading a description of a colour you have never seen is a poor substitute for seeing it.
Language arose as a way of transmitting knowledge about the state of the world between the people that live in it, and it evolves as the world changes over time. Severing this link, by analysing text as a static standalone artifact, leads to problems with grounding of concepts and effectively eliminates exploratory mapping of the language.
In contrast, when humans communicate they generally do so in connection to the environment and interactively in both directions, which provides the necessary grounding of concepts but also an immediate feedback on every utterance. The importance of feedback for human language learners was shown by \citeauthor{sachs1981language} in \cite{sachs1981language}. This paper describes the case of Jim, a hearing child that was brought up by two deaf parents and had to learn to speak from watching television without any supervision or feedback. These circumstances severely delayed his acquisition of language and he did not learn to speak properly until after intervention from the outside. This indicates that not even humans can learn to master human language without the help of guided exploration via synthesis and feedback, i.e. social interactions.

In this paper we investigate if a grounded language can emerge by letting two agents invent their own language to solve a shared problem using Deep Recurrent Q-Networks (DRQN).
More precisely, we let two agents play the game of \emph{Guess Who?} which forces the agents to come up with grounded words that represent characteristics of objects in images in order to win the game.

Recent success in deep reinforcement learning~\cite{mnih-dqn-2015,silver2016mastering} has shown that complex tasks with huge state spaces can be learnt using reinforcement learning (RL). One such task is multi-agent communication. A difficult task due to a states pace which includes all possible conversation states, and non stationary environment which continually change while the agents learn.
In a paper by \citet{NIPS2016_6398} it is shown that agents can learn \emph{continuous} communication using RL, later used to solve tasks that require synchronisation via a global communication channel. Though promising, the continuous nature of this communication makes it very different to natural language where words are discrete symbols.
This deficit was rectified by \citet{foerster2016learning, DBLP:journals/corr/FoersterAFW16} using \emph{Differentiable inter-agent learning} (DIAL) where a method was developed to solve puzzles in a multi-agent setting where the agents were allowed to communicate by sending one-bit messages. Though their communication is still continuous during training, to enable gradient propagation, they regularise the model to promote discrete solutions and discretize during testing. The model we employ in this paper is similar to DIAL, but differs in some key areas. (1) Instead of communicating using bits we generalise the model to handle orthogonal messages of arbitrary dimension to enable vocabularies of arbitrary size, (2) by gradually increasing the noise on the communication channel we ensure that the agents learn a symbolic language but with less negative impact on the convergence rate during training and with a positive impact on the learning capacity (for more on the model see Section~\ref{sec:model}) and (3) in our model no parameters are shared between the agents since this is more reasonable from a human perspective.
In \cite{lazaridou2016multi} a first step towards grounding was taken by letting the agents play a game involving images. The main difference to our paper consist of (1) that we learn the representations of the images while \citeauthor{lazaridou2016multi} use a pretrained classifier trained to recognise the categories of objects present in the images, and (2) that we are able to train the agents to hold a multi-step dialogue in contrast to a one step one symbol question.

The main contributions of this paper include:
\begin{itemize}
    \item An \emph{end-to-end} trainable multiple-agent reinforcement learning model that learns a near optimal strategy for playing \emph{Guess who?} without sharing any parameters between agents and with no predefined communications protocol.
    \item We are able to show that the agents learn to hold a conversation where knowledge from previous time steps are used to guide further dialogue.
    \item An analysis of the invented language that shows how the words are grounded in the concepts visible in the images.
    \item Experiments that show how increasing levels of noise in the communication channel leads to an improved training speed and learning capacity (compared to constant noise) while retaining the ability to learn discrete symbols.
    \item Finally, we generalise DIAL to use orthogonal messages of arbitrary dimension, to more closely resemble human language which encompasses hundreds of thousands of words, and we show that this improves the performance of the system as well as makes it more interpretable.
\end{itemize}

\section{Background}
\label{Background}
The results in this paper rely mainly on the theory of reinforcement learning, deep Q-networks, and the concept of differentiable inter-agent learning~\cite{foerster2016learning}. We introduce these topics briefly below.

\subsection{Reinforcement learning}
\label{RL}
In the traditional \emph{single-agent} reinforcement learning (RL) setting,  an agent observes the current state $s_t \in \mathcal{S}$ at each time step $t$, takes action $u_t \in \mathcal{U}$ according to some policy $\pi$, receives the reward $r_t$, and transitions, according to some probability distribution depending on the current state and action, to a new state $s_{t+1} \in \mathcal{S}$. Value function approaches to RL seek to estimate the value of different policies, typically the expected return, in order to select a good one. The return of a policy is the sum of (discounted) future rewards $R_t = \sum_{\tau = t}^\infty \gamma^{t-\tau}r_\tau$, where $\gamma\in [0, 1]$ is a discount factor that trades-off the importance of immediate and future rewards. For a specific policy $\pi$, the value of a state-action pair is defined as $Q^\pi(s, u) = \mathbb{E}[R_t \,|\, s_t = s, u_t = u]$. The optimal value function $Q^*(s, u) = \max_\pi Q^\pi(s, u)$ is called the \emph{Q-value} of the state-action pair $(s, t)$ and obeys the Bellman optimality equation $Q^*(s, t) = \mathbb{E}[r + \gamma \max_{u'} Q^*(s', u')\, | \,s, u ]$~\cite{sutton1998reinforcement}. An agent that employs the optimal strategy is guaranteed to achieve the highest expected discounted return.
Reinforcement learning can in general be extended to cooperative multi-agent settings where each agent $a$ observes a global state $s_t$,  selects individual actions $u^a_t$, and then receives a team reward $r_t$, shared among all agents. When agents can only partially observe the environment, the global state $s_t$ is hidden and the agents only receive observations $o_t$ that are correlated with the state $s_t$.

\subsection{Deep Q-Networks}
\label{DQN}
 The space of state-action pairs is, in many applications, so large that storing and updating the Q values for each state-action pair is computationally intractable. One solution for this dimensionality problem is to employ the concept of \emph{Deep-Q-Networks} (DQN) (for a more thorough description, see \cite{mnih-dqn-2015}). The idea of DQN is to represent the Q-function by using a neural network parameterised by $\theta$, i.e., to find $Q(s, u; \theta)$ which approximates the value $Q^*(s, u)$ for all state-action pairs. The network is optimised by minimising the loss function $\mathcal{L}_i(\theta_i) = \mathbb{E}[(y_i^{DQN} - Q(s, u; \theta_i)^2]$, at iteration $i$, with  $y_i^{DQN} = r + \gamma\max_{u'}Q(s', u'; \theta_i^{-})$, where $\theta^{-}$ are the parameters of a target network which is fixed for a number of iterations. The actions chosen during the training of the network are determined by an $\epsilon$-greedy policy that selects the action that maximises the Q-value for the current state with probability $1 - \epsilon$ and chooses an action randomly with probability $\epsilon$. When agents only have partial observability, \citeauthor{DBLP:journals/corr/HausknechtS15} (2015) propose to use an approach called \emph{Deep Recurrent Q-Networks} (DRQN) where, instead of approximating the Q-values with a feed-forward network, they approximate the Q-values with a recurrent neural network that can maintain an internal state which aggregates the observations over time. This is modelled by adding an input $h_{t-1}$ to the network that represents the hidden state of the network.

\subsection{Differentiable inter-agent learning}
\label{sec:DIAL}
\citeauthor{foerster2016learning} (2016a) introduce the idea of centralised training but decentralised execution, i.e., the agents are  trained together but evaluated separately. They introduce the concept of \emph{Differentiable Inter-Agent Learning} (DIAL) where messages are allowed to be continuous during training, but need to be discrete during evaluation, which allows gradients to propagate between the agents through the messages in training. This gives the agents more feedback and thus reduces the amount of learning required through trial-and-error. To reduce the discretisation error that could occur from the discretisation of messages in the evaluation phase the messages are processed by a \emph{dicretise/regularise unit} (DRU). During centralised learning the DRU regularises the messages according to DRU($m_t^a$) = Logistic($N(m_t^a, \sigma^2)$), where $m_t^a$ is the message of agent $a$ in time step $t$, and during decentralised execution DRU($m_t^a)=\mathbf{1}\{m_t^a>0\}$. The added noise $\sigma$ during the training phase pushes the messages towards the ends of the logistic function and therefore forces the agents to send almost discrete messages. The change of setup also requires a slight change in how the model is trained; the loss with respect to the $Q$-function is the same as in the DQN case but the gradient term for $m_t^a$ is the error backpropagated through the message from the recipient to the sender. More algorithm details can be found in supplementary materials to \cite{foerster2016learning}. Using DIAL instead of traditional independent $Q$-learning techniques is shown to be beneficial, both in terms of learning speed and in terms of ability to learn. 

\section{Guess Who?}
\label{sect:GuessWho_intro}
The task considered in this paper is a version of the popular guessing game \emph{Guess Who?}. In Guess Who? two players are each assigned one character from a set of 24 characters. Each character is represented by an image and the goal of the game is to deduce which of the 24 characters the other player has. The players take turns asking questions based on the visual appearance of the characters. The questions have to be answered truthfully with \emph{yes} or \emph{no} until one player knows which character the other player was assigned.

The version of Guess Who? that we consider in this paper is a slight variation on the traditional game. One important difference is that we remove the competitive element and have one agent be the \emph{asking-agent} (agent $a = 1$) and the other the \emph{answering-agent} (agent $a = 2$). This means that one agent specialises in asking questions and the other in answering them, and the two of them collaborate with the objective of solving the task. Another difference is that instead of finding the correct character amongst the full set of characters a subset is sampled and given as observation input to the asking agent; in our experiments we used a subset of two or four images. The answering agent observes only the correct image, taken from the subset observed by the asking-agent.

\begin{figure}
    \begin{center}
        \scalebox{0.87}{\includegraphics{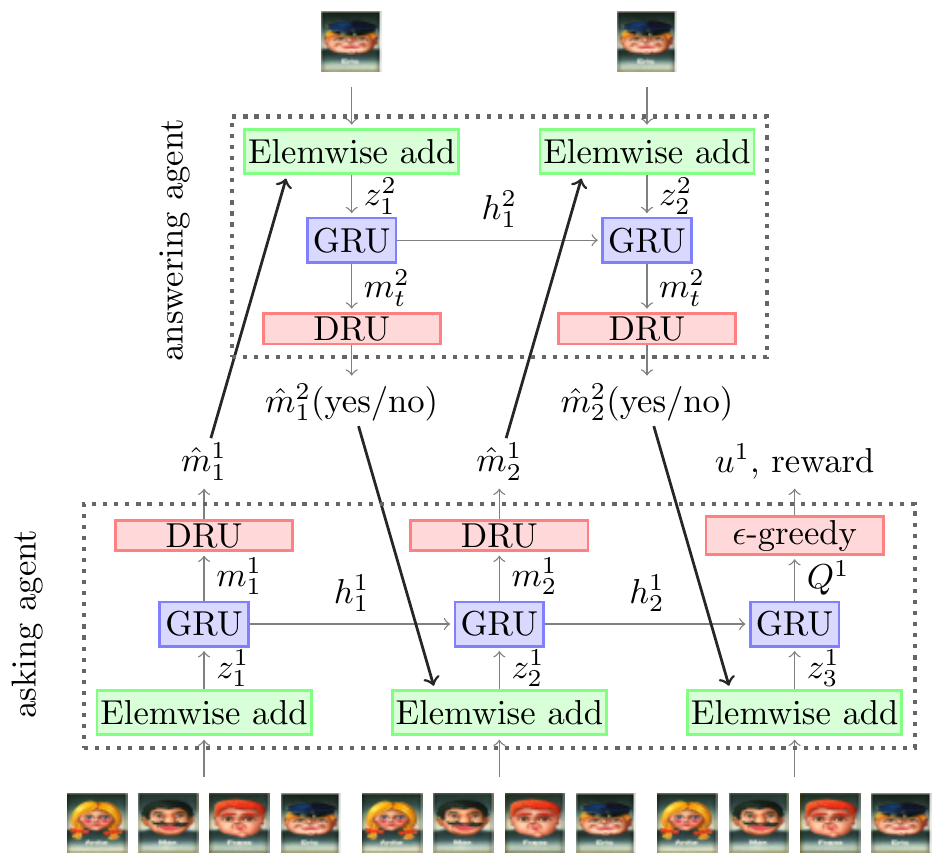}}
    \end{center}
    \caption{Schematic illustration of our version of the Guess Who? game.}
    \label{fig:schematic}
\end{figure}

In Figure \ref{fig:schematic} an example of our version of Guess Who? is illustrated. The asking-agent has four images and the answering-agent has the (correct) image which is the third image of the asking-agent. The asking-agent sends its first question ($m_1^1$) and receives the answer ($m_1^2$). Then another round of question and answer occurs and finally the asking-agent guesses which of the images the answering-agent holds.

In each time step, the agents take turns sending and receiving messages. This means that in each step either the answering-agent receives a question or the asking-agent gets an answer. The questions that the asking-agent may send are limited to a discrete vocabulary of words and the answers the answering-agent can send are limited to two words (\emph{yes} or \emph{no}). Instead of playing the game with a variable amount of time steps a fixed number of steps is played. For the case when the asking-agent observes two images, the agent is only allowed to ask one question ($m_1^1$) and receive one answer ($m_1^2$), and for the case when the asking-agent observes four images, two questions ($m_1^1, m_2^1$) and two answers ($m_1^2$, $m_2^2$) are allowed. After all answers have been received the asking-agent has to guess which of the images the answering-agent has. The guess is considered as an action taken by the asking-agent and is represented by $u^1\in \{1, \dots, n\}$, where $n$ is the number of images the agent holds. If $u^1$ is equal to the index of the character that the asking-agent has, a reward of 1 is given to both agents, otherwise they receive 0.

When the number of question-answer rounds is limited it is not always possible to win each game since only a fixed number of words are available and only yes or no answers are permitted. If if one question is selected from two words it is possible to partition the set of images into four parts. With four words the set can be partitioned in 16 ways. If there are fewer partitions than the size of the set of images this means that several images will be in the same partition, giving the same answers to the questions posed and therefore making them indistinguishable for the model. This implies that the maximum average reward for a game where the asking-agent holds two images from a total of 24 classes (as in Guess Who?) and two different words are allowed is only $0.89$ (for derivations see the supplementary material to this paper). The equivalent score in a four image game with two rounds of questions would be $0.71$ (for derivations see the supplementary material to this paper). However,  since our agents have recurrent networks it is possible that the questions in each round have different meanings (otherwise there would be no point in asking the same question in the second round as in the first round) such that the set can be partitioned further, implying that the maximum average reward is higher than $0.71$.

In order to evaluate the proposed framework on a larger set of images and more realistic images, we also use the \emph{CelebA} dataset \cite{liu2015faceattributes} which includes 202,599 face images of 10,177 unique celebrity identities. The images in the dataset cover large pose variations and background clutter. Examples of images can be seen in Table \ref{tab:celeba}.

\begin{table}[]
    \centering        \caption{Examples of images in the CelebA dataset}

    \renewcommand\tabcolsep{0pt}
    \begin{tabular}{c c c}
       \adjustimage{width=0.2\linewidth,valign=m}{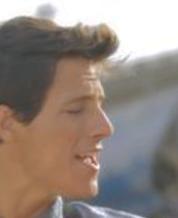} &
       \adjustimage{width=0.2\linewidth,valign=m}{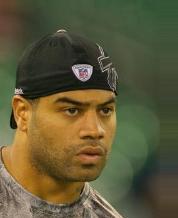}  & \adjustimage{width=0.2\linewidth,valign=m}{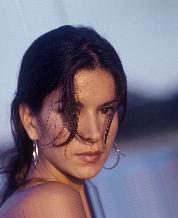}\\
    \end{tabular}
    \label{tab:celeba}
\end{table}

\section{Model}
\label{sec:model}

\begin{figure}[t]
    \centering
    \includegraphics[width=\linewidth]{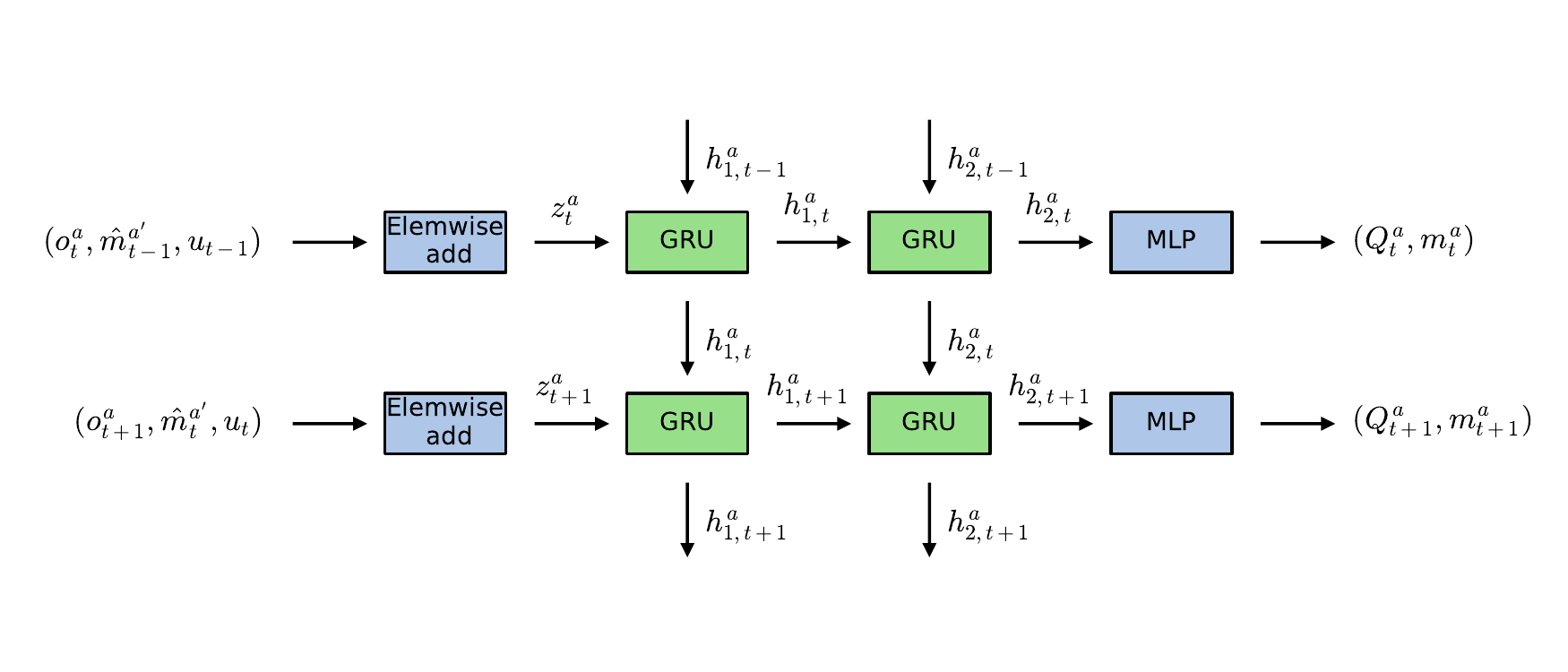}
    \caption{Architecture of the model. The time dimension (question-answer rounds in the game) of the RNN goes from top to bottom. The green boxes illustrate the internal state of the network.}
    \label{fig:model}
\end{figure}

The architecture of the model is presented in this section, a schematic illustration of the model is shown in Figure \ref{fig:schematic} and a more detailed overview of each agent is shown in Figure \ref{fig:model}. Each agent  ($a = 1, 2$) consists of a \emph{recurrent neural network} (RNN) that is rolled out for $T$ steps. The agents have an input network, a 2-layer gated recurrent unit (GRU) \cite{DBLP:conf/emnlp/ChoMGBBSB14} and an output network. The input network takes the incoming information images ($o_t^a$), message ($\hat{m}_{t-1}^{a'}$) and the agent's previous action ($u_{t-1}^a$), embeds them separately into embeddings using Multi-Layer Perceptrons (MLP) before adding the embeddings element-wise to create the joint embedding $z_t^a$. The embedding $z_t^a$ and the incoming state $h^a_{(1,2),t-1}$ is passed through the 2-layer GRU generating new states $h^a_{(1,2),t-1}$. The state of the top layer is the output embedding $h^a_{2,t-1}$. The output embedding is passed through the output network generating $Q_a,  m_t^a$. $Q_a(u)$ is used to generate action $u_t^a$ using an $\epsilon$-greedy policy. The message $m_t^a$ is passed through a variant of DRU as described in section \ref{sec:DIAL} to generate a one-hot encoding using $\hat{m}^t_a=$DRU($m_t^a$) = Softmax($N(m_t^a, \sigma^2_{episode})$) in the training case where $N(\mu,\sigma^2)$ is the $|m|$ dimensional normal distribution. During evaluation $\hat{m}^t_a(i)=\{1$ if $i=\argmax(m), 0$ otherwise$\}$ is used instead. The parameters for creating the embeddings and for generating the output $Q^a$ and $m_t^a$ can be found in Table \ref{tab:modelnetworks}. \\

In our model we utilise messages in a \emph{one-hot encoding} which are then passed to the other agent for the next time step. While using a one-hot encoding does have disadvantages in terms of scalability (since a binary encoding of length $n$ gives $2^n$ different possible messages instead of only $n$ in the one-hot encoding) we found that it gave improved results and makes it easier to study the underlying effect behind each message which is clouded by the varying closeness between messages that occurs in binary encoding.

\begin{table}[t]
    \centering
    \caption{Desription of the networks that generate the embeddings for the input and generate the output}
    \begin{tabular}{l l r}
    \toprule
    & Network type & Layer sizes \\
     \midrule
   Embeddings & & \\
    \cmidrule(r){1-1}
                     $o_t^a$ &  2-layer MLP     & 128, 256  \\
                     $u_{t-1}^a$ &  Lookup & 256  \\
                      $\hat{m}_{t-1}^{a'}$ &  MLP  &  256  \\

    Recurrent network && \\
     \cmidrule(r){1-1}
     & 2-layer GRU & 256,256 \\
    Output & & \\
    \cmidrule(r){1-1}
                          $m$ and $Q$ &  2-layer MLP & 256, $|u| + |m|$  \\
    \bottomrule
    \end{tabular}
    \label{tab:modelnetworks}
\end{table}

Batch normalization (see, \cite{DBLP:conf/icml/IoffeS15}) is performed in the MLP for the image embedding and on the regularised incoming messages $\hat{m}_{t-1}^{a'}$. During testing, non-stochastic versions of batch normalization is used which implies that running averages of values observed during training are used instead of those from the batch.

Updates of the parameters of the network are performed as described in section \ref{sec:DIAL} with more details available in supplementary materials to \cite{foerster2016learning}.

\section{Increasing noise}
Inspired by curriculum learning proposed by \citet{Bengio:2009:CL:1553374.1553380}, who showed that gradually increasing difficulty of tasks leads to improved learning, we introduce the usage of increasing noise to DIAL. We found that using a fixed value for the noise $\sigma$ led to an unwelcome trade off. If $\sigma$ is too large there is a risk of the model never learning anything. If $\sigma$ is too small the training error (where continuous messages are allowed) is excellent but the testing error is bad due to the model over-encoding information in the messages leading to a large discretisation error.  Our solution to this is to allow the noise to linearly increase over the epochs. This enables the model to learn quickly in the beginning but to punish the over-encoding of information more and more as the training progresses. This can be compared to human speech where a person who is beginning to learn a new language requires an almost noise free surrounding to understand a message while a proficient speaker would have no problem understanding a message in a more noisy environment such as listening to the radio or having a conversation by a busy road.

\section{Experiments}
To validate our ideas we run a series of experiments on our version of Guess Who? to see how well it performs but also what the model learns.  Further, we evaluate the effectiveness of increasing noise by comparing it to experiments with non-increasing noise.

For each episode we randomly select two or four different images from the pool of 24 Guess Who? images (which are down-sampled to $32\times32$ pixels). The model is trained with $\sigma$ increasing linearly from 0.1 to 1. We evaluate the model performance with two, four, or eight different words available to the asking agent in a one-hot embedding, and the answers yes/no available to the answering agent. The game goes on for three time steps (i.e., one question, one answer, and one guess) for two images and five time steps (i.e, two questions and two answers) for four images.  This experiment is also replicated for the CelebA dataset creating a pool of 202,599 images, here we only study the case with four images and for eight and sixteen different words.

Finally, we sample the messages between agents to understand the properties of the communication. This is to observe if the agents adapt their messages to the history of the communication, creating a conversation rather than individual messages. We also want to see if the meaning of words are constant along the conversation or are dependent on their context. We also compare our model with a model where the incoming state of the GRU of the answering agent gets zeroed out (and also the previous action), removing the possibility of these depending effects that depend on the history of the conversation, to see how this affects the performance of the model.

The experiments use an $\epsilon$-greedy policy with $\epsilon = 0.05$, a discount factor $\gamma=1$, and each epoch performs parallel episodes using a batch size of 32. The target network $\theta^-$ is updated every 100 epochs. Unless stated otherwise, we let $\sigma$ increase linearly from 0.1 to 1; this keeps the average $\sigma$ close to what was found to work on a task with similar tasks in \cite{foerster2016learning}, where they used $\sigma=0.5$. The optimization is done by RMSProp~\cite{tieleman2012lecture} with a learning rate of $5\times 10^{-4}$. Each experiment is run multiple times and the results averaged. The code for the experiments is available online at \href{https://github.com/emiliojorge/Inventing-a-Grounded-Language}{github.com/emiliojorge/Inventing-a-Grounded-Language}.

\section{Results}

\subsection{Performance}
\begin{figure}
\centering
 \includegraphics[width=0.9\linewidth]{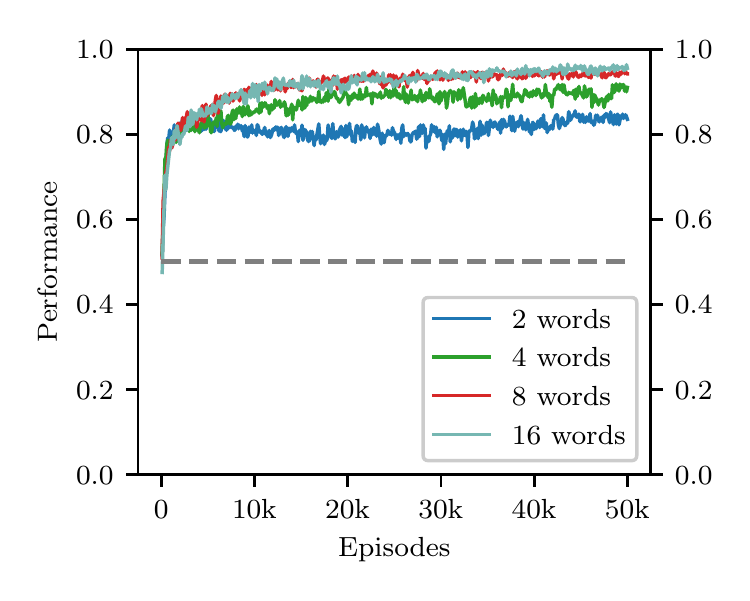}
 \caption{Performance of the model on Guess Who? images when the asking-agent has two images, a vocabulary of two, four, eight or sixteen different words and one round of question-answer is performed. The results are averaged over five runs. The dashed grey lines represents the baseline performance where the asking-agent guesses randomly. The performance of the model is ordered from the highest score with the largest vocabulary to the lowest score with the smallest vocabulary.}
  \label{fig:real2}
\end{figure}

\begin{figure}
  \centering
  \includegraphics[width=0.9\linewidth]{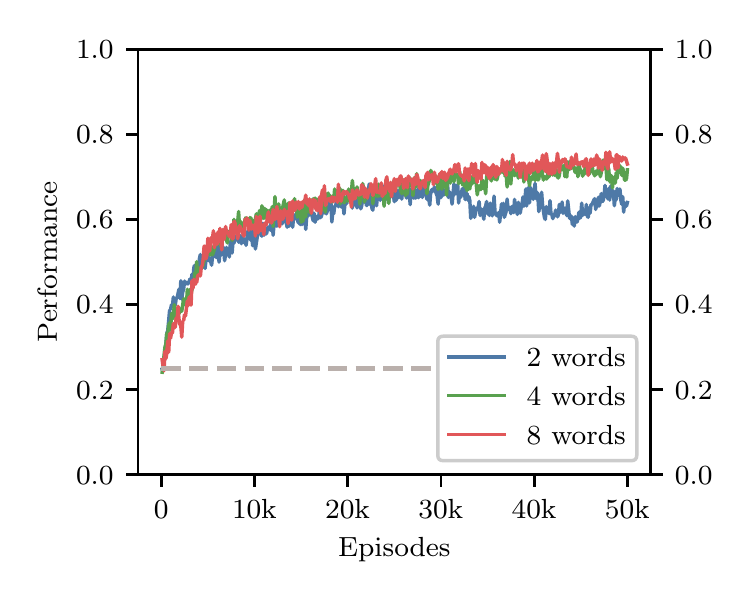}
  \caption{Performance of the model on Guess Who? images when the asking-agent has four images, a vocabulary of two, four, or eight different words and two rounds of question-answer is performed. The results are averages of five runs. The dashed grey lines represents the baseline performance where the asking-agent guesses randomly. The performance of the model is ordered from the highest score with the largest vocabulary to the lowest score with the smallest vocabulary.}
  \label{fig:real4}
\end{figure}

In Figure \ref{fig:real2} the average performance (over five runs) is illustrated in the case where the asking-agent sees two images Here it is clear that more words are crucial for better performance on two images. The score for two words is close to the theoretical bound of $0.89$ for two words (as discussed in Section \ref{sect:GuessWho_intro}) but is surpassed when using four and eight words.

The performance for four images averaged over five runs can be seen in Figure \ref{fig:real4}.
It can be seen that the performance obtained on four images is quite similar independently of how many questions are available. 
\begin{figure}
    \centering
    \includegraphics[width=0.9\linewidth]{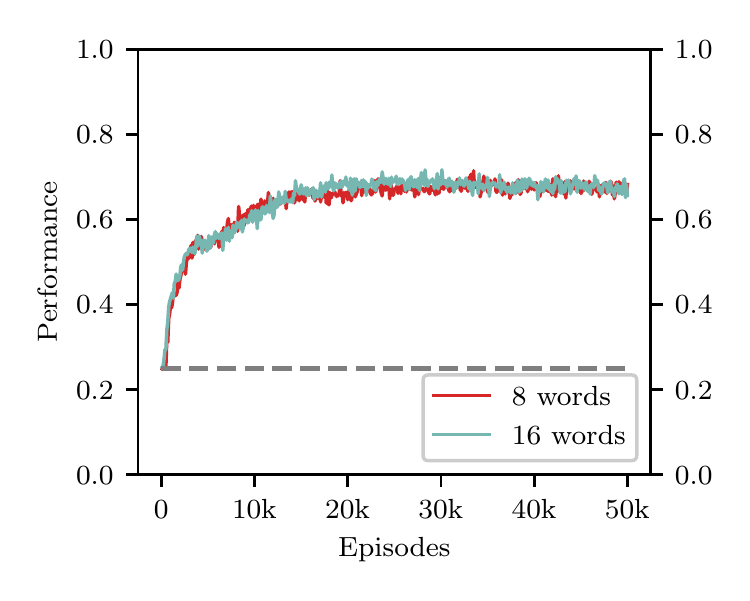}
    \caption{Performance of the model on images from the CelebA dataset when the asking-agent has four images, two rounds of question-answer are performed and with a vocabulary of eight and sixteen words available. The dashed grey lines represents the baseline performance where the asking-agent guesses randomly.}
    \label{fig:celeb4}
\end{figure}
The results of the experiments on the CelebA dataset can be seen in Figure \ref{fig:celeb4}. This shows that performance is similar even when the set of images is very large. This indicates that the performance of the model is not dependent on the networks memorising the images but rather generalises to combinations of images it is very unlikely to have seen before. This idea is also supported by a model trained on $90\%$ of the images and tested on the remaining $10\%$ obtaining very similar scores when evaluated on the different data.
\subsection{Understanding the questions}
\begin{figure}
    \centering
    \includegraphics[width=0.85\linewidth]{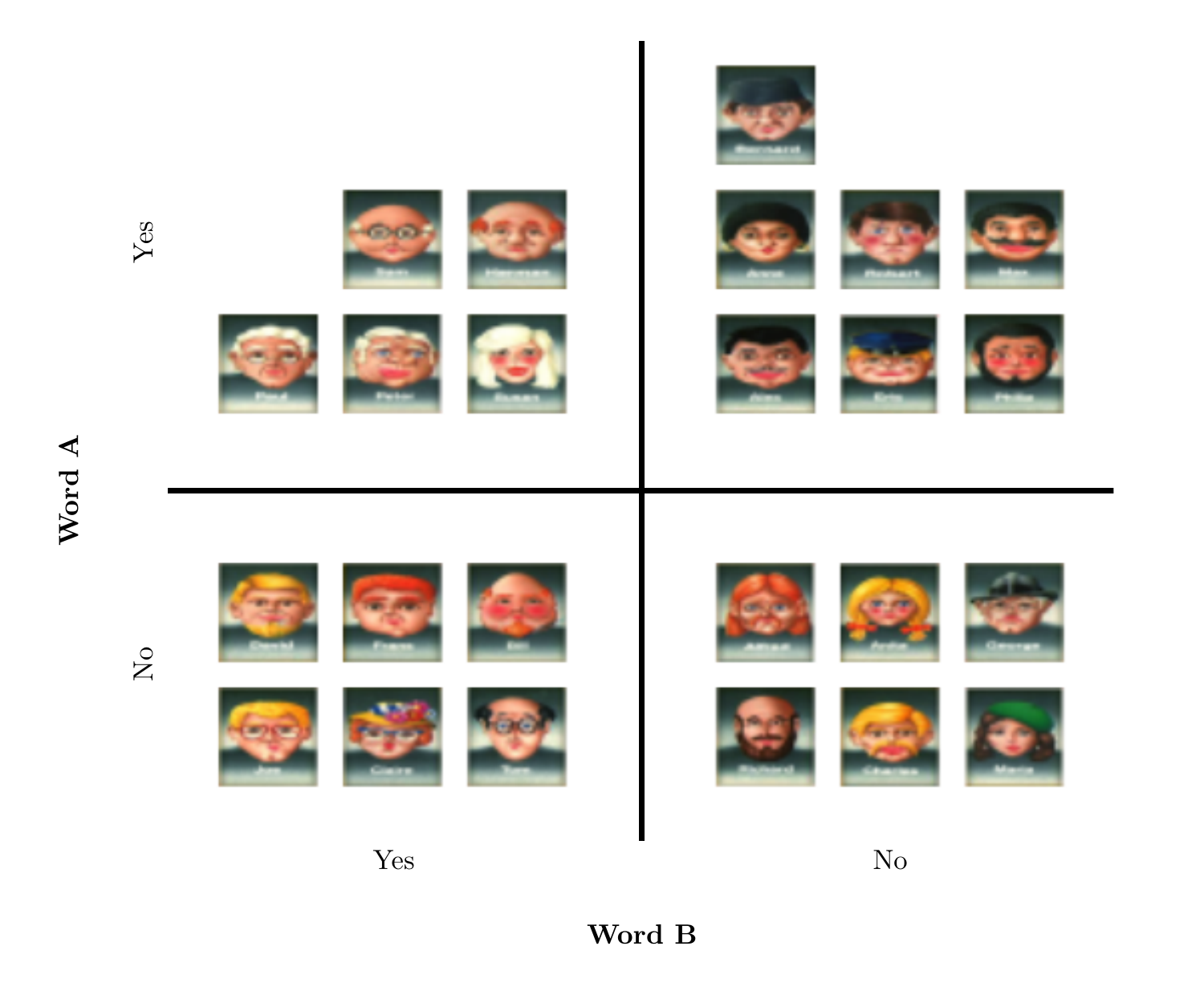}
    \caption{A partition of the Guess Who? images defined by the answers the answering-agent gives for each question (for the case when the vocabulary is of size two).}
    \label{fig:division}
\end{figure}

To better understand the nature of the questions that the asking-agent asks we sample a few episodes from a model with two images and two available questions to see how the games play out, i.e., what the message protocol between the agents is depending on the images they see. The message protocol is illustrated in Table \ref{tab:sample_games}. In the table it can be seen that the asking-agents questions depend on what images it has, the answers it gets from the answering-agent, and its interpretation of the answer (its guess). One can see that in some cases it receives the same answer even when the answering-agent has different images. This leads to the answering-agent sometimes making a guess on the wrong image and receiving zero reward.
\begin{table}[]
    \caption{Learned message protocols between the asking-agent and the answering-agent depending on the images the agents see. The vocabulary is of size two and the asking agent has two images and is allowed one question/answer}
    \centering
        \begin{tabular}{cccccc}
        \toprule
        & & \multicolumn{2}{c}{Messages}\\
        \cmidrule(r){3-4}
        Asker & Answerer & Q & A & Guess & Reward\\
        \midrule
        \adjustimage{width=0.07\linewidth,valign=m}{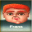}\adjustimage{width=0.07\linewidth,valign=m}{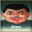} & \adjustimage{width=0.07\linewidth,valign=m}{guess_who/8.png} & B & yes & \adjustimage{width=0.07\linewidth,valign=m}{guess_who/8.png} & 1 \\
        \adjustimage{width=0.07\linewidth,valign=m}{guess_who/8.png}\adjustimage{width=0.07\linewidth,valign=m}{guess_who/1.png} & \adjustimage{width=0.07\linewidth,valign=m}{guess_who/1.png} & B & no & \adjustimage{width=0.07\linewidth,valign=m}{guess_who/1.png} & 1 \\
        \adjustimage{width=0.07\linewidth,valign=m}{guess_who/1.png}\adjustimage{width=0.07\linewidth,valign=m}{guess_who/8.png} & \adjustimage{width=0.07\linewidth,valign=m}{guess_who/8.png} & B & yes & \adjustimage{width=0.07\linewidth,valign=m}{guess_who/8.png} & 1 \\
        \adjustimage{width=0.07\linewidth,valign=m}{guess_who/1.png}\adjustimage{width=0.07\linewidth,valign=m}{guess_who/8.png} & \adjustimage{width=0.07\linewidth,valign=m}{guess_who/1.png} & B & no & \adjustimage{width=0.07\linewidth,valign=m}{guess_who/1.png} & 1 \\
        \midrule
        \adjustimage{width=0.07\linewidth,valign=m}{guess_who/8.png}\adjustimage{width=0.07\linewidth,valign=m}{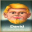} & \adjustimage{width=0.07\linewidth,valign=m}{guess_who/8.png} & A & no & \adjustimage{width=0.07\linewidth,valign=m}{guess_who/2.png} & 0 \\
        \adjustimage{width=0.07\linewidth,valign=m}{guess_who/8.png}\adjustimage{width=0.07\linewidth,valign=m}{guess_who/2.png} & \adjustimage{width=0.07\linewidth,valign=m}{guess_who/2.png} & A & no & \adjustimage{width=0.07\linewidth,valign=m}{guess_who/2.png} & 1 \\
        \adjustimage{width=0.07\linewidth,valign=m}{guess_who/2.png}\adjustimage{width=0.07\linewidth,valign=m}{guess_who/8.png} & \adjustimage{width=0.07\linewidth,valign=m}{guess_who/8.png} & A & no & \adjustimage{width=0.07\linewidth,valign=m}{guess_who/2.png} & 0 \\
        \adjustimage{width=0.07\linewidth,valign=m}{guess_who/2.png}\adjustimage{width=0.07\linewidth,valign=m}{guess_who/8.png} & \adjustimage{width=0.07\linewidth,valign=m}{guess_who/2.png} & A & no & \adjustimage{width=0.07\linewidth,valign=m}{guess_who/2.png} & 1 \\

        \bottomrule
        \end{tabular}
    \label{tab:sample_games}
\end{table}
The answers to each question for the different images is illustrated in Figure \ref{fig:division}. This shows that the two images that it has trouble separating (see the ones yielding zero reward in Table \ref{tab:sample_games}) have the same answers to both words (\emph{No} to word A and \emph{Yes} to word B), as such it is impossible for the asking-agent to distinguish between the two. This illustrates the problem with a limited vocabulary which leads to only being able to partition the set in a few ways. Judging from the appearances of the images, it seems like word A could be interpreted as something along the lines of \emph{Is the top of his/her head very light coloured or very darkly coloured?}.

\begin{figure}
    \centering
    \includegraphics[width=0.85\linewidth]{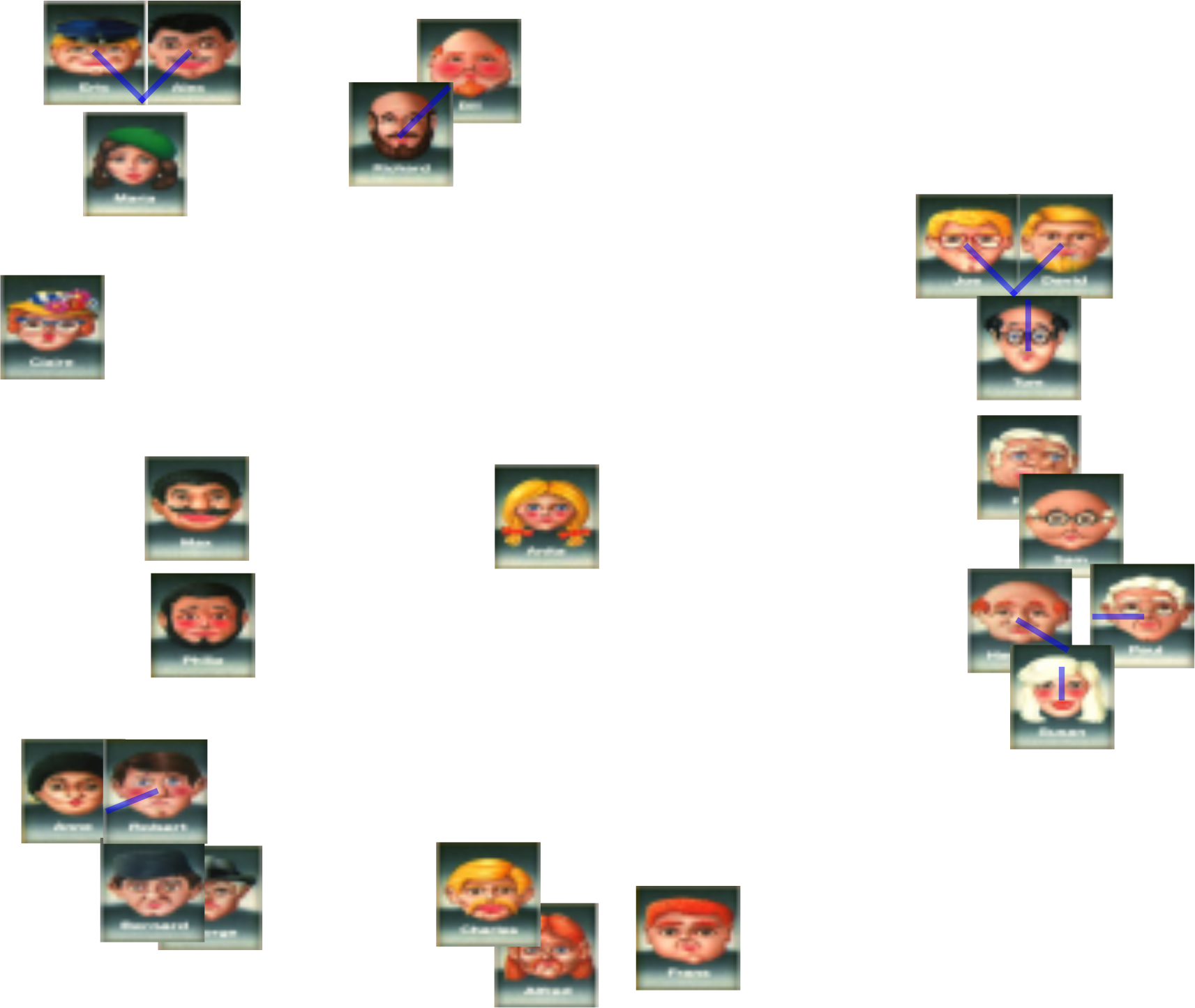}
    \caption{A t-SNE embedding of the images using answers to questions as a distance measure. Blue lines indicate the correct position for images that have been shifted to reduce visual cluttering.}
    \label{fig:embedding}
\end{figure}

The division, as shown in Figure \ref{fig:division}, was reproduced twelve times. How often images have the same answers to questions can be used as a distance measure to describe the relation between images. Using this distance measure we then project the images to a two-dimensional space using t-Distributed Stochastic Neighbor Embedding (t-SNE) \cite{maaten2008visualizing} which can be seen in Figure \ref{fig:embedding}. This projection also shows that similar images appear have similar answers to questions to a large degree.

\subsection{Interactive conversation}
\begin{figure}
    \centering
    \includegraphics[width=0.9\linewidth]{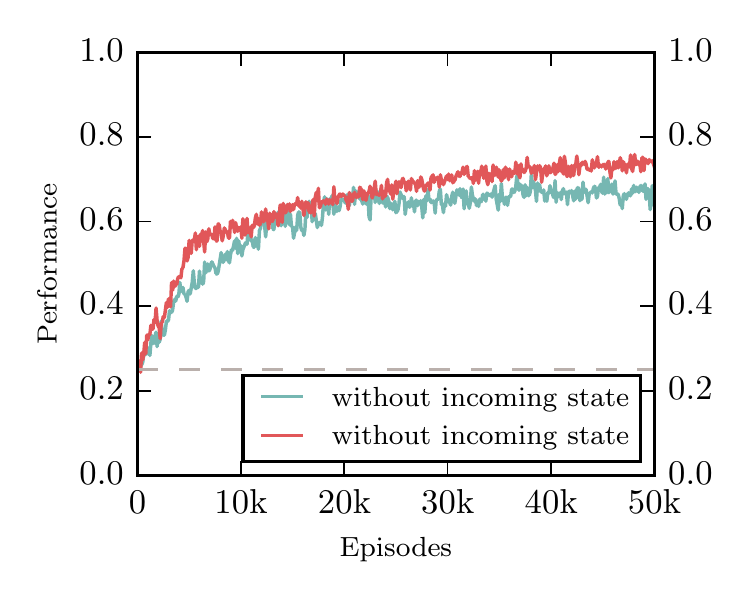}
    \caption{Comparison of the performance of model with and without an incoming state for the answering agent. The comparison is performed on images from the Guess Who? dataset with a vocabulary of eight words, a subset of four images and two question/answers. The dashed grey lines represents the baseline performance where the asking-agent guesses randomly. The results are average performance over 6 runs.}
    \label{fig:zero_out}
\end{figure}
The comparison seen in Figure \ref{fig:zero_out} highlights the benefits of having an interactive communication. When the answering agent has a memory of previous communication it allows the communication to create homographs. Homographs are words that are spelt the same but have different meaning depending on the context. One example of a homograph is the word rock, which can both refer to a style of music but also to a stone. Since the agents have a memory of the conversation in the GRUs it allows the agents to adapt the meaning to allow words to have different significance in the second round of question answers depending on the history of the communication. Since we use a multiple step communication with agents using GRUs it is possible for the agents to create an interactive conversation where the messages sent depend on the internal state of each agent. This allows the agents to adapt the messages to send depending on the history of the communication such that a greater variety of communication is possible. By studying what message the asking agents sends as the second question in a model trained on CelebA with a vocabulary of 8 words, we find that keeping everything else the same (images, first question), the second questions differs in $93\%$ of situations depending on what the answer to the first question was. This shows that the agents create some kind of conversation adapting to the other agent instead of just using sending messages independently. This also makes sense, instead of limiting to $8\times8=64$ possible question pairs created by the ability to form homographs it instead becomes $8\times2\times8=128$.
\subsection{Inreasing noise}
\begin{figure}
    \centering
    \includegraphics[width=0.9\linewidth]{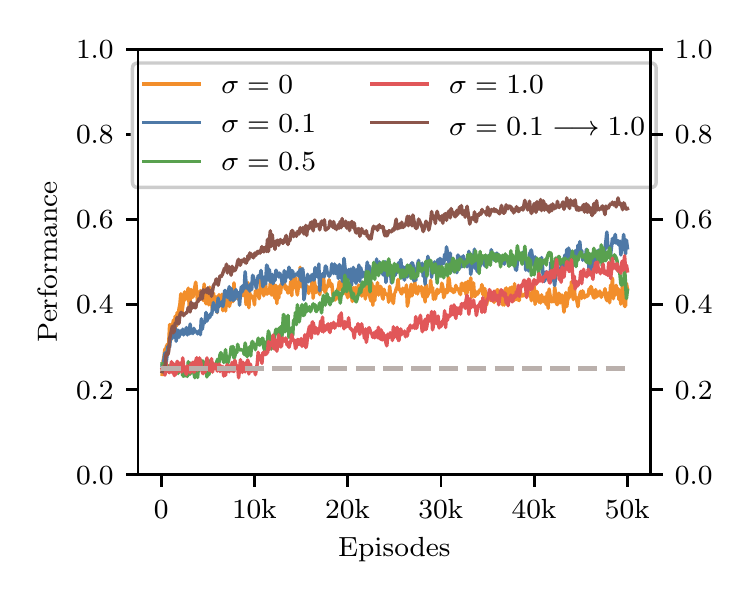}
    \caption{The performance of the model for different noise levels ($\sigma \in \{0,0.1, 0.5, 1.0\}$), and for $\sigma$ varying linearly from $0.1$ to $1.0$.) The dashed grey line represents the baseline performance where the asking-agent guesses randomly. The results are averaged over twelve runs for the increasing $\sigma$ and five runs for constant $\sigma$. The performance of the model with increasing $\sigma$ outperforms the models with constant $\sigma$.}
    \label{fig:noise}
\end{figure}

In order to evaluate the effectiveness of increasing the noise $\sigma$ during training we compare results on Guess Who? using four images and with two available words and an increasing $\sigma$ with a constant $\sigma = 0.5$ (which was used in \cite{foerster2016learning}) but also with $\sigma\in\{0,0.1,1\}$.
In Figure \ref{fig:noise} it is clearly visible that there is a significant advantage in using variable noise compared to constant noise. The model learns both faster and achieves better performance compared to the models trained with constant noise. This supports the idea that the communication is grounded in visual aspects of the images.
\section{Conclusions} 
In this paper we have shown that a DRQN can learn to play Guess Who?. Further, by careful analysis of the results we could establish a grounding connection between the words used by the agents and characteristics of objects in images from the game. We also show that the agents create an interactive conversation where the communication adapts to what has previously been said in the communication. Finally, we showed that regulating the level of noise in the communications channel could have a large impact on the training speed as well as the final performance of the system.

\subsubsection*{Acknowledgments}
The authors would like to acknowledge the project \emph{Towards a knowledge-based culturomics} supported by a framework grant from the Swedish Research Council (2012--2016; dnr 2012-5738).
The authors would also like to thank Jakob Foerster and Yannis M. Assael for the insightful discussions.

\bibliography{ref.bib}

\newpage
\appendix
\onecolumn
\section{Derivations for optimal scoring with two words}

\label{app:calculations}
When trying to correctly differentiate between two images from a pool of 24 images using one out of two available question the probability of guessing the correct image can be seen as the following: 
Let image A be the correct image and B be the incorrect image from the pool.
\begin{equation}
\label{eq:2images}
\begin{aligned}
    P(\text{Correct}) & =   P(\text{Correct}|\text{Separable}) \cdot P(\text{Separable}) \\
    & +   P(\text{Correct}|\text{Not separable}) \cdot P(\text{Not separable}). \\
\end{aligned}
\end{equation}
Images are separable if the images A and B are not in the same partition of images divided up by the questions. Since a partition that divides the set up equally is optimal this means that the 24 images are divided up such that there are 6 images in each partition and $P(\text{Separable})=1-\nicefrac{5}{23}=\nicefrac{18}{23}$. If they are inseparable asking agent will have to perform a random guess. Equation \ref{eq:2images} gives $P(\text{Correct})=1 \cdot \nicefrac{18}{23} + \nicefrac{1}{2} \cdot \nicefrac{5}{23} \approx 0.89$.

For the case with four images it is slightly different. The following calculations assume that two questions are asked but that those are the only two questions available such that the two questions available in the first question phase are the same as in the second phase. This means that the optimal procedure will be to ask one of them in the first phase and the other in the second phase. In this case the calculations are as above but take into consideration that there may be multiple images in the correct partition. This gives
\begin{equation*}
\begin{aligned}
    P(\text{Correct}) & =  P(\text{Correct}|\text{Separable}) \cdot P(\text{Separable}) \\
    & +    P(\text{Correct}|\text{Not separable w/one}) \cdot P(\text{Not separable w/one}) \\
    & +   P(\text{Correct}|\text{Not separable w/two}) \cdot P(\text{Not separable w/two}) \\
    & +    P(\text{Correct}|\text{Not separable w/three}) \cdot P(\text{Not separable w/three})\\
      &   =  \frac{\frac{1}{1}\binom{5}{0} \binom{18}{3} + \frac{1}{2}\binom{5}{1} \binom{18}{2} + \frac{1}{3} \binom{5}{2} \binom{18}{1} + \frac{1}{4}\binom{5}{3} \binom{18}{0}} {\binom{23}{3}} \approx 0.71.
\end{aligned}
\end{equation*}

\end{document}